\documentclass[]{article}
\makeatletter\if@twocolumn\PassOptionsToPackage{switch}{lineno}\else\fi\makeatother

\usepackage{amsmath,tabulary,graphicx,times,caption,fancyhdr,amssymb,amsfonts,amstext,amsbsy}
\usepackage[utf8]{inputenc}
\usepackage[T1]{fontenc}
\usepackage{authblk}
\usepackage{booktabs}
\usepackage{makecell}
\usepackage{graphicx}

\usepackage[paperheight=10in,paperwidth=6.5in,margin=2cm,headsep=.5cm,headheight=1.5cm,top=2.5cm]{geometry}
\renewenvironment{abstract} {\vspace*{-1pc}\trivlist\item[]\leftskip\oupIndent\hrulefill\par\vskip4pt\noindent\textbf{\abstractname}\mbox{\null}\\}{\par\noindent\hrulefill\endtrivlist} 
 \date{}
\usepackage{url,multirow,morefloats,floatflt,cancel,tfrupee}
\makeatletter

\AtBeginDocument{\@ifpackageloaded{textcomp}{}{\usepackage{textcomp}}}
\makeatother
\usepackage{colortbl}
\usepackage{xcolor}
\usepackage{pifont}
\usepackage[nointegrals]{wasysym}
\makeatletter

\def\mcWidth#1{\csname TY@F#1\endcsname+\tabcolsep}

\def\cAlignHack{\rightskip\@flushglue\leftskip\@flushglue\parindent\z@\parfillskip\z@skip}
\def\rAlignHack{\rightskip\z@skip\leftskip\@flushglue \parindent\z@\parfillskip\z@skip}

\@ifundefined{etal}{}{}

\usepackage{ifxetex}
\ifxetex\else\if@twocolumn\@ifpackageloaded{stfloats}{}{\usepackage{dblfloatfix}}\fi\fi

\AtBeginDocument{
\expandafter\ifx\csname eqalign\endcsname\relax
\def\eqalign#1{\null\vcenter{\def\\{\cr}\openup\jot\m@th
  \ialign{\strut$\displaystyle{##}$\hfil&$\displaystyle{{}##}$\hfil
      \crcr#1\crcr}}\,}
\fi
}

\AtBeginDocument{%
  \@ifpackageloaded{endfloat}%
   {\renewcommand\efloat@iwrite[1]{\immediate\expandafter\protected@write\csname efloat@post#1\endcsname{}}}{\newif\ifefloat@tables}%
}%

\def\BreakURLText#1{\@tfor\brk@tempa:=#1\do{\brk@tempa\hskip0pt}}
\let\lt=<
\let\gt=>
\def\processVert{\ifmmode|\else\textbar\fi}

\@ifundefined{subparagraph}{
\def\subparagraph{\@startsection{paragraph}{5}{2\parindent}{0ex plus 0.1ex minus 0.1ex}%
{0ex}{\normalfont\small\itshape}}%
}{}

\newcommand\role[1]{\unskip}
\newcommand\aucollab[1]{\unskip}
  
\@ifundefined{tsGraphicsScaleX}{\gdef\tsGraphicsScaleX{1}}{}
\@ifundefined{tsGraphicsScaleY}{\gdef\tsGraphicsScaleY{.9}}{}
\def\checkGraphicsWidth{\ifdim\Gin@nat@width>\linewidth
	\tsGraphicsScaleX\linewidth\else\Gin@nat@width\fi}

\def\checkGraphicsHeight{\ifdim\Gin@nat@height>.9\textheight
	\tsGraphicsScaleY\textheight\else\Gin@nat@height\fi}

\def\fixFloatSize#1{}
\let\ts@includegraphics\includegraphics

\def\inlinegraphic[#1]#2{{\edef\@tempa{#1}\edef\baseline@shift{\ifx\@tempa\@empty0\else#1\fi}\edef\tempZ{\the\numexpr(\numexpr(\baseline@shift*\f@size/100))}\protect\raisebox{\tempZ pt}{\ts@includegraphics{#2}}}}

\AtBeginDocument{\def\includegraphics{\@ifnextchar[{\ts@includegraphics}{\ts@includegraphics[width=\checkGraphicsWidth,height=\checkGraphicsHeight,keepaspectratio]}}}

\DeclareMathAlphabet{\mathpzc}{OT1}{pzc}{m}{it}

\def\URL#1#2{\@ifundefined{href}{#2}{\href{#1}{#2}}}

\def\UrlOrds{\do\*\do\-\do\~\do\'\do\"\do\-}%
\g@addto@macro{\UrlBreaks}{\UrlOrds}

\edef\fntEncoding{\f@encoding}

\makeatother

\newif\ifmultipleabstract\multipleabstractfalse%
\usepackage[noindentafter]{titlesec}
\def\NormalBaseline{\def\baselinestretch{1.1}}

\titleformat{\section}[hang]{\NormalBaseline\filright\large\fontsize{12}{15}\bfseries\boldmath}
{\large\thesection.}
{10pt}
{\noindent}
[]
\titleformat{\subsection}[hang]{\NormalBaseline\filright\fontsize{11}{13}\bfseries\itshape\boldmath}
{\thesubsection.}
{10pt}
{}
[]
\titleformat{\subsubsection}[hang]{\NormalBaseline\filright\fontsize{10}{12}\bfseries\itshape\boldmath}
{\thesubsubsection.}
{10pt}
{}
[]
\titleformat{\paragraph}[runin]{\NormalBaseline\filright\itshape}
{\theparagraph.}
{10pt}
{}
[]
\titleformat{\subparagraph}[runin]{\NormalBaseline\filright\itshape}
{\thesubparagraph.}
{10pt}
{}
[]

\titlespacing{\section}{0pt}{1.5\baselineskip}{.2\baselineskip}  
\titlespacing{\subsection}{0pt}{1\baselineskip}{.2\baselineskip}  
\titlespacing{\subsubsection}{0pt}{1.5\baselineskip}{.2\baselineskip}  
\titlespacing{\paragraph}{0pt}{.5\baselineskip}{10pt}  
\titlespacing{\subparagraph}{0pt}{.5\baselineskip}{10pt}

\makeatletter\def\oupIndent{1pt}
\def\author#1{\gdef\@author{\hskip-\dimexpr(\tabcolsep)\hskip\oupIndent\parbox{\dimexpr\textwidth-\oupIndent}{\centering\bfseries#1}}}
\def\title#1{\gdef\@title{\centering\bfseries\ifx\@articleType\@empty\else\@articleType\\\fi#1}}
\let\@articleType\@empty \def\articletype#1{\gdef\@articleType{{\normalfont\itshape#1}}}
\fancypagestyle{headings}{\fancyhf{}\fancyhead[C]{\RunningHead\hspace*{1pc}}\fancyhead[R]{\thepage}}
\makeatother
\usepackage[authoryear]{natbib}

\makeatletter
\@ifpackageloaded{hyperref}{\def\alreadybookmark{}}{\usepackage[hidelinks]{hyperref}}
\@ifpackageloaded{bookmark}{}{\usepackage{bookmark}}
\makeatother
\ifdefined\alreadybookmark\def\pdfbookmark[#1]#2#3{}\fi
\usepackage{float}

\begin{document}
\pdfbookmark[title]{WDR FACE: The First Database For Studying Face Detection In Wide Dynamic Range }{title}
\nocite{*}

\title{WDR FACE: The First Database For Studying Face Detection In Wide Dynamic Range}
\author{Ziyi Liu\textsuperscript{1}\thanks{E-mail: ziyi.liu1@ucalgary.ca}{ },
            Jie Yang\textsuperscript{2}\thanks{Corresponding author. E-mail: yangjie@westlake.edu.cn}~,
            Mengchen Lin\textsuperscript{1},
            Kenneth Kam Fai Lai\textsuperscript{3},
            Svetlana Yanushkevich\textsuperscript{3} and
            Orly Yadid-Pecht\textsuperscript{1}~\\[-3pt]\normalsize\normalfont\itshape 
~\\\textsuperscript{1}{I2Sense lab\unskip, University of Calgary\unskip, Calgary\unskip, Canada}
~\\\textsuperscript{2}{Westlake University\unskip, Hangzhou\unskip, China}
~\\\textsuperscript{3}{Biometric Technologies Laboratory\unskip,
Schulich School of Engineering\unskip,  University of Calgary\unskip, Calgary\unskip, Canada}}
\def\RunningHead{{\mbox{}}}

\def\RunningHead{{WDR FACE: The First Database For Studying Face Detection In Wide Dynamic Range}}

\maketitle

\begin{abstract}
Currently, face detection approaches focus on facial information by varying specific parameters including pose, occlusion, lighting, background, race, and gender. These studies only utilized the information obtained from low dynamic range images, however, face detection in wide dynamic range (WDR) scenes has received little attention.  To our knowledge, there is no publicly available WDR database for face detection research.  To facilitate and support future face detection research in the WDR field, we propose the first WDR database for face detection, called WDR FACE, which contains a total of 398 16-bit megapixel grayscale wide dynamic range images collected from 29 subjects. These WDR images (WDRIs) were taken in eight specific WDR scenes. The dynamic range of 90\% images surpasses 60,000:1, and that of 70\% images exceeds 65,000:1.  Furthermore, we show the effect of different face detection procedures on the WDRIs in our database. This is done with 25 different tone mapping operators and five different face detectors.  We provide preliminary experimental results of face detection on this unique WDR database.
\end{abstract}\def\keywordstitle{Keywords}

    
\section{Introduction}
\label{sect:intro}  
In recent years, wide dynamic range (WDR) imaging has become prevalent in applications such as surveillance, computer vision, video enhancement, etc. \citep{905417:20518622, 905417:20517558}. Face detection is the initial step of various facial applications such as face recognition \citep{905417:20517561}, verification \citep{905417:20517560}, and face alignment \citep{905417:20517559}, which is usually performed on 8-bit based images and requires a certain degree of image quality \citep{905417:20518595}. It is possible that a computer vision algorithm fails to detect a face correctly \citep{905417:20518626}, if the face is caught in a bright or dim region. Despite this fact, existing research concentrates on detecting faces over low dynamic range (LDR) images, which often expose the face detection issue under extreme environment. The facial information of LDR images may suffer from saturated pixel values, manifesting overexposure.  To avoid pixel saturation in LDR imaging, the most common way is to reduce the exposure time while photographing. This often leads to capturing fewer photons in the lower end of the dynamic range, causing under-exposure in some dark regions. In this case, the facial details of the dark regions will be not only very limited, but also masked by noise.  Either way, it negatively impacts the result of face detection.
\bgroup
\fixFloatSize{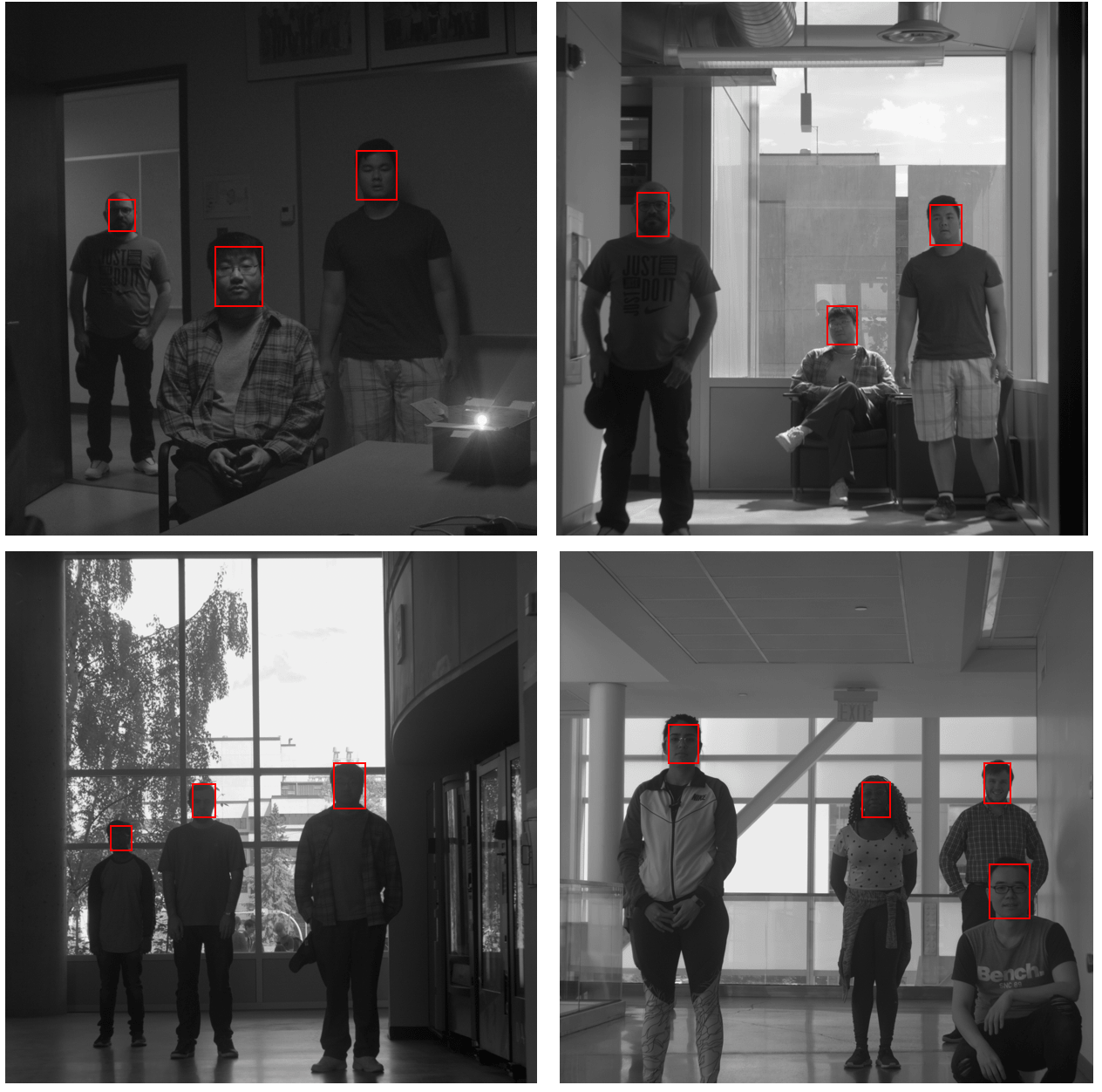}
\begin{figure*}[!htbp]
\centering \makeatletter\IfFileExists{images/1.png}{\includegraphics{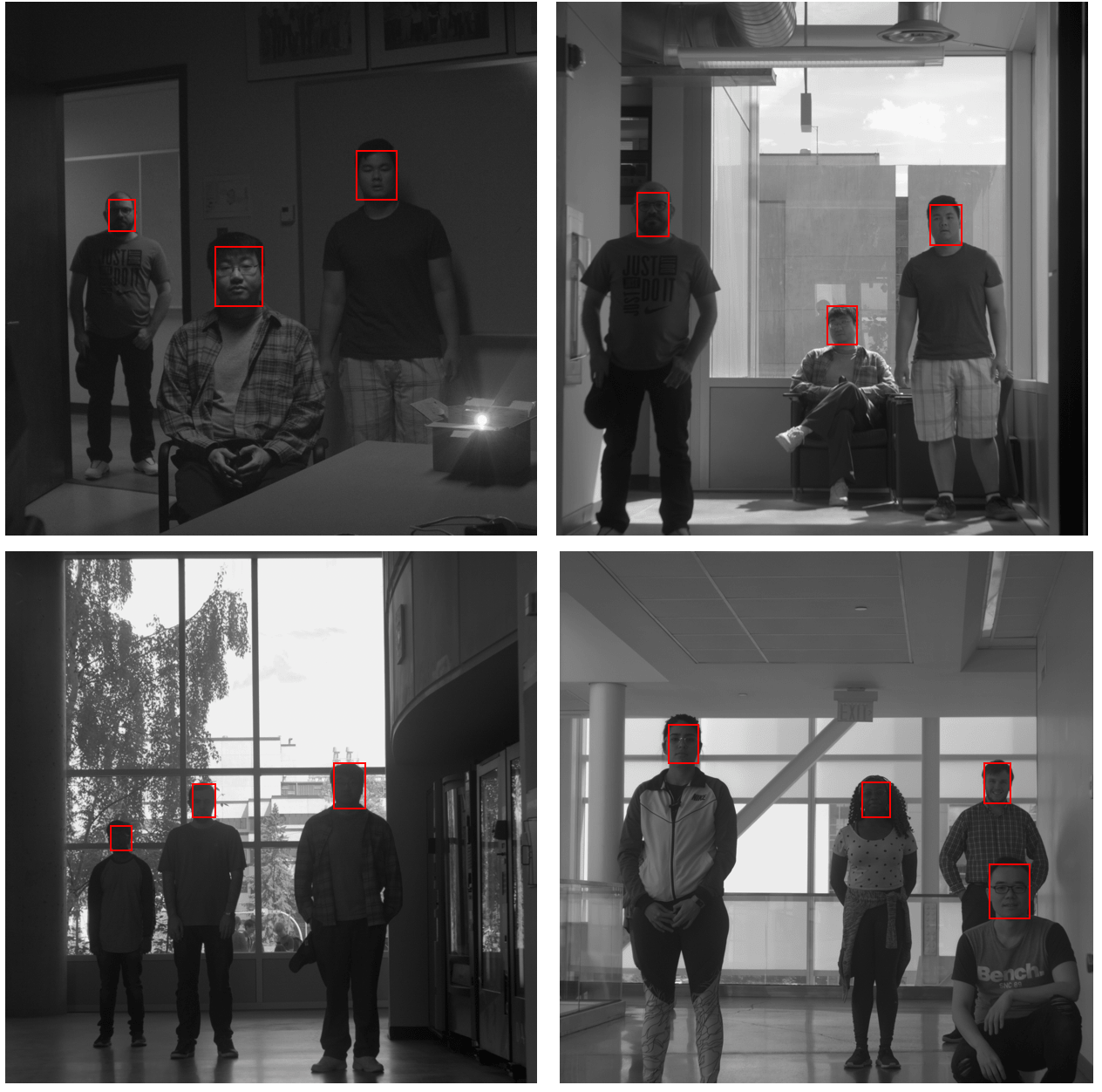}}{}
\makeatother 
\caption{{Example of captured WDR images from WDR FACE database. The red bounding boxes indicate the detected faces. The box coordinates are provided for those images.}}
\label{f-cecf4c97ef25}
\end{figure*}
\egroup

In real life, we often encounter the WDR scenes that exceed the dynamic range capacity of conventional digital cameras, increasing the risk of losing facial details. Fortunately, WDRIs are likely to be able to assist face detectors in achieving better results \citep{905417:20518626}. An algorithmic solution to obtain high illuminance images is to apply multi-exposure fusion (MEF) to a sequence of multi-exposure LDR images, known as exposure bracketing, that is usually captured by a conventional LDR digital camera \citep{905417:20518619}. Image bracketing is one of the automatic functions of most modern digital cameras and mobile cameras \citep{905417:20518620}.  However, the resulting images often suffer from different fusion artifacts, such as ghosting, due to the object movement in WDR scenes, and the jittering of the camera.  Despite the effective and outstanding performance of image fusion approaches, they often fail to generate high-quality images\citep{905417:20518611}. In theory, the best way to acquire WDRIs is to exclude the post algorithmic processes and directly obtain the images via a custom device with a WDR imaging sensor. Unfortunately, expensive and inconvenient WDR equipment discourages researchers from using this approach.  Presently, there is no published WDR database for face detection research.

To support the study in this aspect, we introduce the first WDR database for face detection research. The database contains 29 subjects for a total of 398 16-bit WDRIs. All images were taken by an Andor Zyla PLUS 4.2 megapixel scientific CMOS (sCMOS) camera (Figure. \ref{f-cecf4c97ef25}).  The actual dynamic range of 398 images is between $2^{15}$ (32,768:1) and $2^{16}$ (65,536:1). To diversify our database, we have selected 8 WDR scenes for taking WDRIs. One of the scenes is an artificially designed WDR scene (using a single-point light source to create high illuminance regions in a relatively dark room). The WDR of the other seven scenes is based on ambient light. These scenes are not suitable for shooting with a conventional digital camera.   Owing to the extreme WDR of each scene, the images are especially prone to produce under/over-exposed regions, resulting in very limited information captured in the WDR scenes \citep{905417:20518626}. Since there is no similar WDR database at present,  we show the effect of face detection on our database.  This is done using 25 different tone mapping operators (TMOs) and five different face detectors (see Figure \ref{f-c48060293e1a}).  In addition to describing the details of our database, we provide preliminary experimental face detection results.

A review of the existing databases, tone mapping operators, and face detectors is given in Section \ref{sect:related_work}. We describe the procedure of the image collection, and the details of the database in Section  \ref{sect:database}. The evaluation methods, including the selection of TMOs and face detectors, evaluation procedure and settings are discussed in Section \ref{sect:eval_method}. Next, we provide face detection results in Section \ref{sect:result}. Finally, conclusions are provided in Section \ref{sect:conclusion}.

\begin{table*}[!t]
\footnotesize
\caption{General comparison of our WDR FACE with some most well-known face databases.}
\label{table:db_stat}
\tabcolsep 20pt 
\begin{tabular*}{\textwidth}{@{\extracolsep{\fill}}lccc}
\toprule
Database & Total Images & \# of faces & Dynamic Range \\ \midrule
FERET & 14,126 & - & 255:1 \\
CMU Multi-PIE & $>75,000$ & - &  255:1 \\
CMU PIE & $>41,000$ & - &  255:1 \\
LFW & 13,233 & - & 255:1 \\
FDDB & 2,845 & 5,171 & 255:1 \\
MALF & 5,250 & 11,931 & 255:1 \\
WIDER FACE & 32,203 & 393,703 & 255:1 \\
PubFig & 58,797 & - & 255:1 \\
IJB-A & 24,327 & 49,759 & 255:1 \\
AFW & 205 & 468 & 255:1 \\
WDR FACE & 398 & 796 & \textbf{32,768:1 to 65,536:1} \\ 
\bottomrule
\end{tabular*}
\end{table*}

\section{Related Work}
\label{sect:related_work} 
In this section, we briefly review the related existing databases, state-of-the-art tone mapping operators and face detection methods.
\bgroup
\fixFloatSize{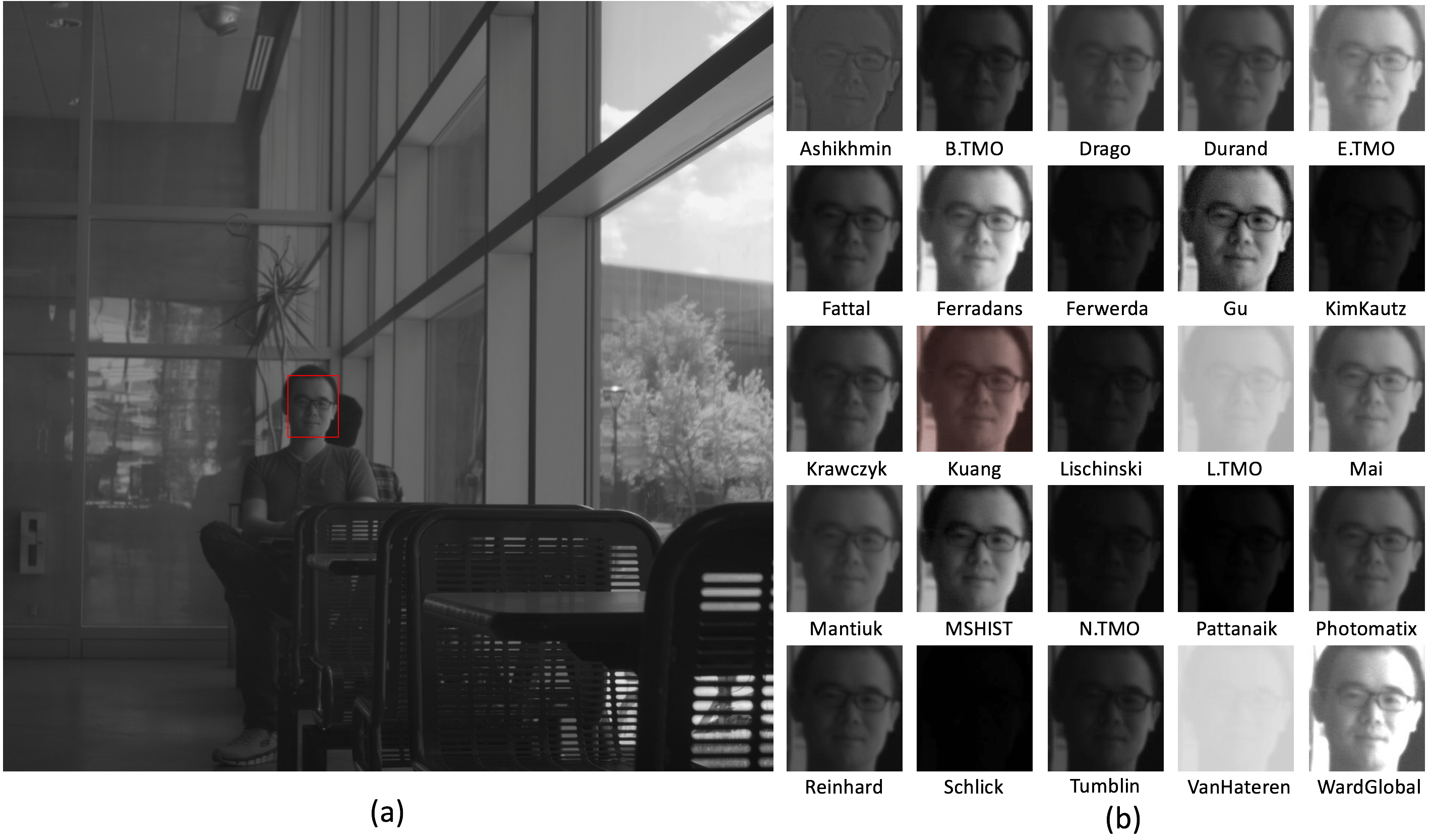}
\begin{figure*}[!htbp]
\centering \makeatletter\IfFileExists{images/2.png}{\includegraphics{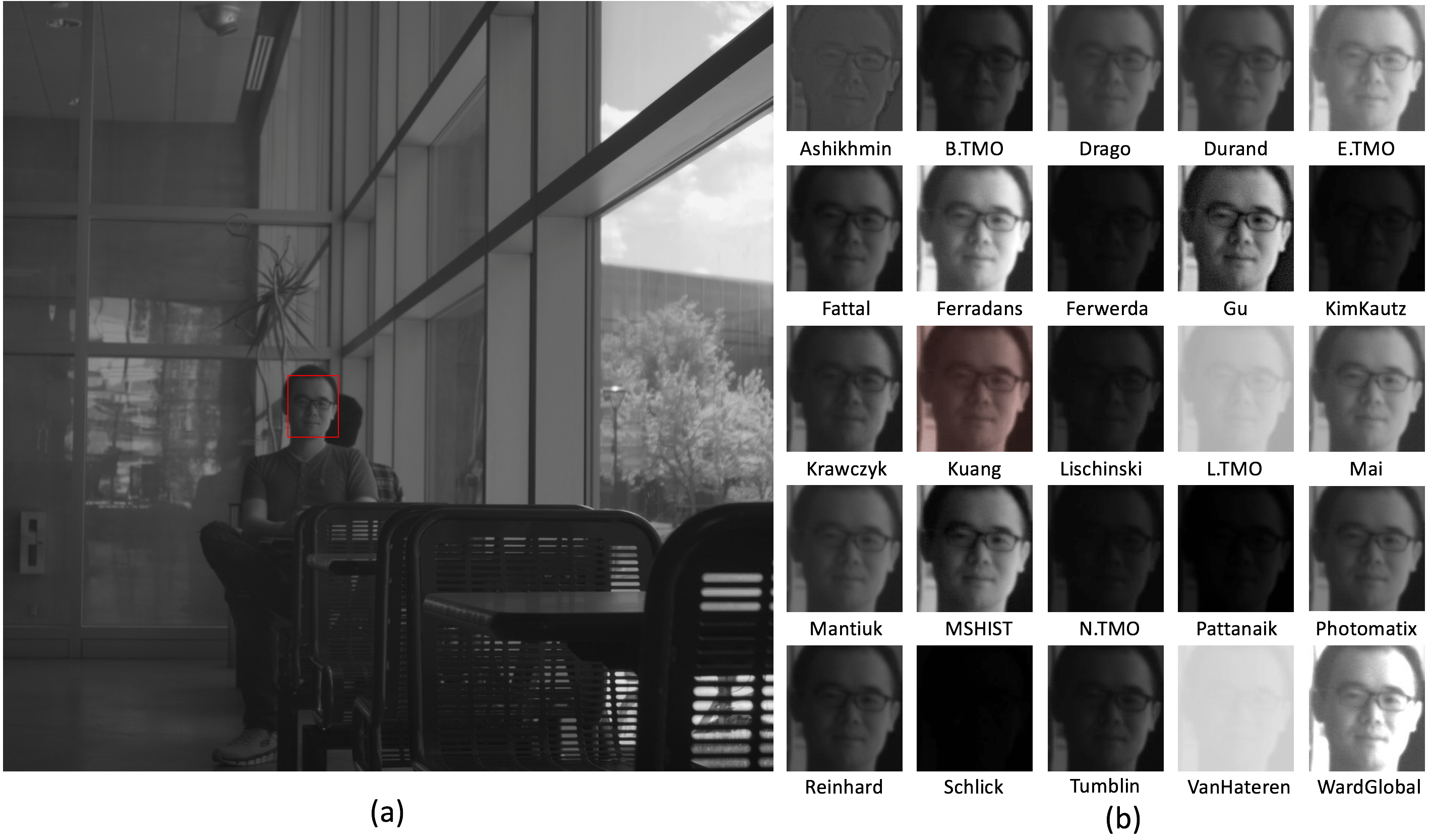}}{}
\makeatother 
\caption{{A sample of tone mapped faces. (a) The LDR representation of a captured WDR image.  (b) The cropped faces of 25 tone mapped images. B., E., N. and L. refer to BestExposureTMO, ExponentialTMO, NormalizeTMO and LogarithmicTMO, respectively.}}
\label{f-c48060293e1a}
\end{figure*}
\egroup

\subsection{Standard Dynamic Range Face Detection Databases}
There are many publicly available face databases online. We summarize the attributes of some of the well-known databases in Table \ref{table:db_stat}.  The images of these face databases, such as AFW \citep{905417:20518610}, MALF \citep{905417:20518602}, LFW \citep{905417:20518603}, FDDB \citep{905417:20518604},  IJB-A \citep{905417:20518608}, WIDER FACE \citep{905417:20518600}, YTF \citep{905417:20518598}, and PubFig \citep{905417:20518601} were drawn from either social networking applications or image search engines. The focus of these databases is thus the diversity of the image content such as a large variation in pose, lighting, occlusions, and a wider selection of ethnicity of subjects. Annotated faces in-the-wild AFW dataset is built using Flickr images. It contains 205 images with 468 labeled faces. MALF, consisting of 5,250 images and 11,931 faces, is the first face detection dataset that supports fine-grained evaluation. LFW benchmark contains 13,233 images collected from 5,749 subjects.  FDDB dataset has the annotations for 5,171 faces in a set of 2,845 images. It is a subset of images from Berg et al.’s dataset \citep{905417:20518597} which were collected from the Yahoo! news website. IJB-A contains 24,327 images and 49,759 faces and it is suitable for the research in both face detection and face recognition. WIDER FACE organized the categories of variability in scale, pose, occlusion, expression, appearance, and illumination.  It consists of 32,203 images that contain 393,703 labeled faces with bounding boxes. YTF has 1,595 subjects with the faces in 3,425 videos. Pubfig contains 58,797 images of celebrities and politicians.
Other databases such as CMU PIE \citep{905417:20518605}, CMU Multi-PIE \citep{905417:20518606}, FERET \citep{905417:20518599} focus on
focus on taking images in a very controlled condition.  It intended to facilitate the study of specific parameters on face identification problems \citep{905417:20518603}. 

Despite the remarkable diversity of the released face databases, they all consist of LDR face images. As far as we are aware, there is no existing face database that concentrates on the dynamic range of images. This motivated us to create the first WDR database for the study of face detection.

\textbf{Synthetic WDR Face Detection Databases:}
Besides the above-mentioned LDR face databases, there are two existing databases \citep{905417:20518595,905417:20518596} which were made of synthetic face images. These images are generated by either MEF algorithms from exposure-bracketed still images, or the TMOs of synthetic WDRIs created by MEF algorithms, depicting groups of people under highly variable lighting conditions. The MEF algorithm recovers the image detail in dark and bright regions such as deep shades and sun flare. Although the WDR scene represented by the synthesized LDR seems to be able to provide visible human faces in the LDR field, the LDR processed by these post-algorithmic algorithms has a lower ability to restore the WDR scene than the high bit depth image \citep{905417:20518595}.  The details lost by illuminance compression from the WDR domain to the LDR domain are also unknown.  Thus, these databases are not suitable for the study of this paper.

\subsection{Tone Mapping Operators}
The research of properly displaying a WDR image on an LDR display has been carried on for decades.  Many researchers have proposed complicated tone mapping techniques to generate LDR representations from WDRIs.  These TMOs can be categorized into global \citep{905417:20518675, 905417:20518638, 905417:20518594, 905417:20518637} and local \citep{905417:20518681, 905417:20518640, 905417:20518639, 905417:20518644, 905417:20518669, 905417:20518647, 905417:20518642} operators.  Global TMO applies its function equally to all pixels regardless of the surrounding of the pixel in an image.  Khan \citep{905417:20518638} used a sensitivity model of the human visual system.  Larson's \citep{905417:20518594}  and  Drago \citep{905417:20518637} proposed histogram-based methods and adaptive logarithmic mapping, respectively.  In general, global TMO applications are relatively simple to implement. However, because of ignoring the local features of the input WDR image, the tone-mapped result is liable to suffer from low brightness and low contrast. Sometimes there is a loss of details.  On the other hand, local TMO takes local statistics into account.  The resulting images often achieve higher contrast and are able to reveal more local details.  The prevalent local operators are, for instance, Fattal's \citep{905417:20518669} gradient domain optimization and Durand's \citep{905417:20518644} bilateral approach. Some edge-preserving filters methods such as Paris \citep{905417:20518647}, Gu \citep{905417:20518646}, and Mantiuk \citep{905417:20518642} also achieve compelling results.  Recently, learning-based approaches such as Rana \citep{905417:20518593} show great potential and competitive performance. Gharbi et al. \citep{905417:20518635} trained coefficients of a locally-affine model to effectively enhance an input image.

\subsection{Face Detectors}
Existing face detection algorithms can be roughly divided into four categories: knowledge-based approach, feature invariant approach, template matching approach, and appearance-based approach \citep{905417:20518592}.  Appearance-based face detection methods draw more attention than others since they adopt machine learning techniques to learn distinctive face landmarks.  The fast and accurate Viola-Jones \citep{905417:20518591} face detector falls into this category.  It extracts large amounts of Haar features and adopts Adaboost \citep{905417:20518572} to train the nodes of a cascade framework.  Li et al.\citep{905417:20518590} use SURF feature \citep{905417:20518571} that is more distinctive with a moderate number.  It outperforms Viola-Jones' framework with less running time and higher performance. King \citep{905417:20518587} proposed Max-Margin Object Detection (MMOD) which learns Histogram of Oriented Gradients (HoG) \citep{905417:20518586} on training images using structural Support Vector Machines (SVMs) to optimize overall sub-windows to detect objects. In addition to these features that can be directly observed from the image, there are high-dimensional features that can be extracted using deep learning.  Garcia et al. \citep{905417:20518588} applied Convolutional Neural Network (CNN) to face detection two years after the Viola-Jones method was published and they also achieved good performance.  Zhang et al. \citep{905417:20518585} proposed the Multitask Cascaded Convolutional Networks (MTCNN) framework that adopted three stages of cascaded structure to jointly solve face detection and alignment. In 2015, Ren et al. \citep{905417:20518584} proposed Faster regional convolutional neural networks (Faster R-CNN) which concentrates on boosting up the speed of the R-CNN framework to operate at 7 fps by replacing selective search with a neural network. More mature methods, such as YOLOv3 \citep{905417:20518589}, demonstrated state-of-the-art performance.

\bgroup
\fixFloatSize{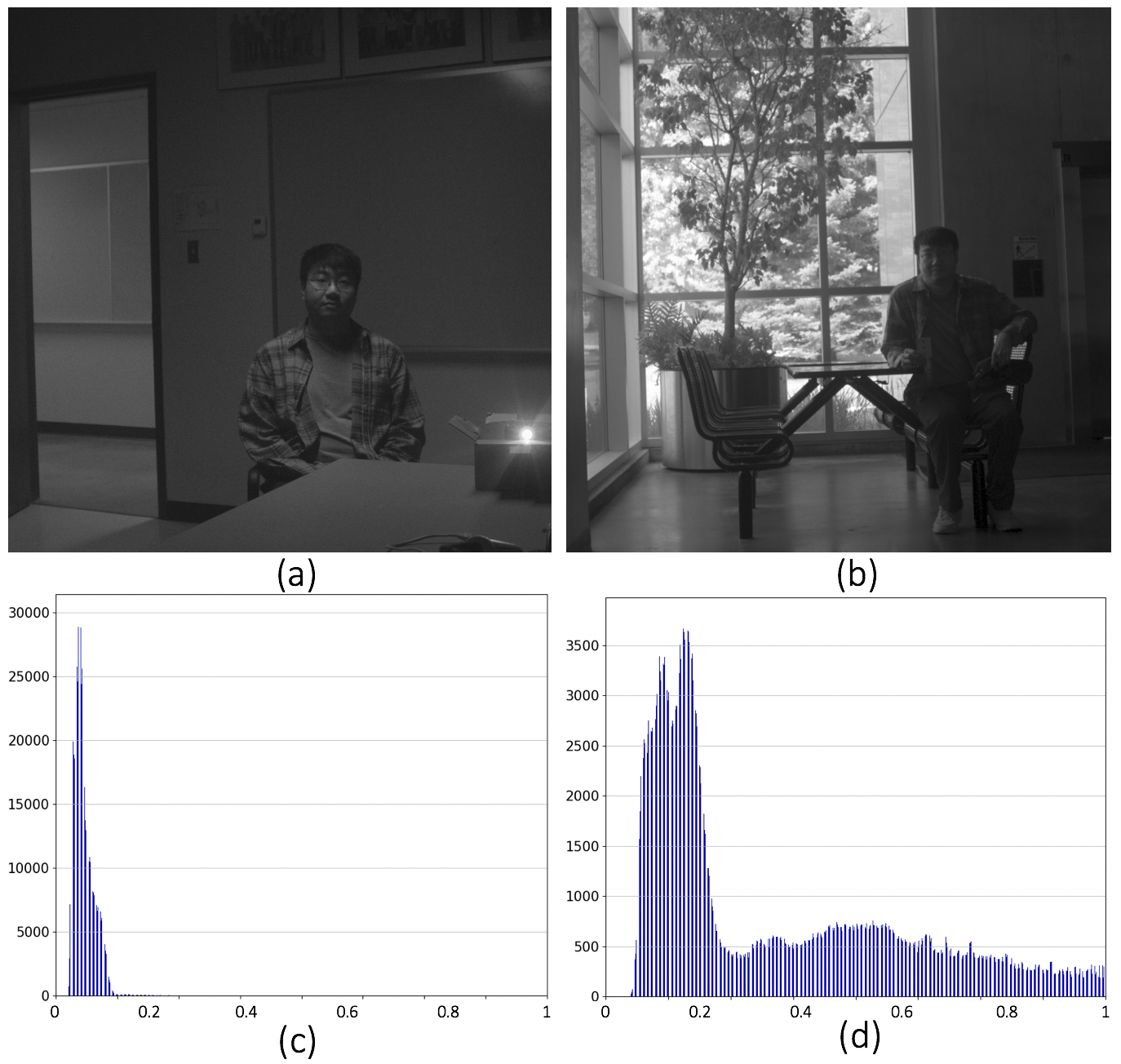}
\begin{figure*}[!htbp]
\centering \makeatletter\IfFileExists{images/3.png}{\includegraphics{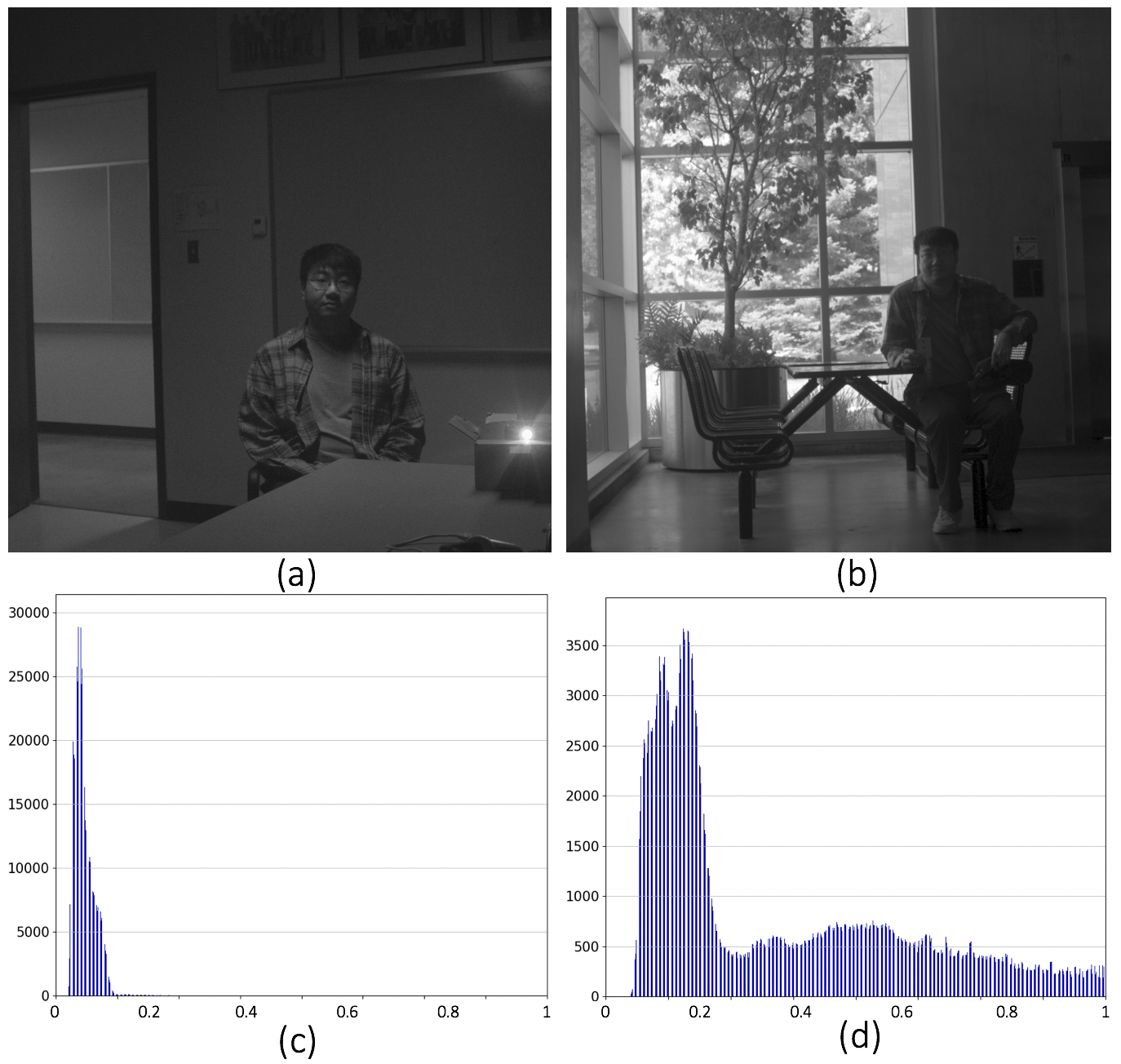}}{}
\makeatother 
\caption{{Demonstration of the histogram of the artificial and a natural scene. (a) The image captured from the artificial WDR Scene. (b) An image captured from one of the natural scenes.  (c) The histogram of (a).  Most pixels' value gathered in the lower end.  (d) The histogram of (b).}}
\label{f-4c0a2e69a07f}
\end{figure*}
\egroup

\section{WDR Face Database}
\label{sect:database} 
\subsection{Data Collection}
\paragraph{Overall.}
To our best knowledge, the WDR Face dataset is the first Wide Dynamic Range dataset applying for face detection task, which is captured on 8 scenes. The images in the dataset are collected following the below three steps:
\begin{itemize}
    \item Select scenes.
    \item Take the photographs.
    \item Clean data.
\end{itemize}
\textbf{Select Scenes.}
We expect to diversify the illuminance distribution of the WDRIs, particularly, the illuminance distribution of the captured faces. We achieve this by manipulating the scene diversity. The complex WDR scenes can be roughly divided into two categories according to the light source: artificial scenes and natural scenes. Artificial scenes (usually indoors) are dominated and constructed lighting systems. For instance, the CMU 3D Room \citep{905417:20518582} could provide extremely controlled lighting conditions, which is able to expedite the study of the impact of changing a specific parameter on the result. Yet it is usually not enough to approximate the conditions in everyday life \citep{905417:20518603}.  In real-world scenarios, illumination is often composed of ambient light with one or two point sources \citep{905417:20518605}.  The distribution of illuminance in natural scenes will be much more complex. We show an example of histogram of illuminance distribution in Figure  
\ref{f-4c0a2e69a07f}.

In order to provide illuminance distribution data in two different categories of WDR scenes, we selected seven natural scenes and set up a simple artificial scene. The artificial WDR scene consists of a flashlight point source in a relatively dark environment, of which the dynamic range could reach up to around 1:65536. The other seven WDR natural scenes were in the daytime containing window to satisfy the wide dynamic range criterion. In addition, these scenes with subject participation all have backlighting. The excessive backlight can cause the subjects' faces to be shown only silhouettes, which increases the difficulty of face detectors to locate the facial landmarks. In order to eliminate the monotonic strong backlighting, we selected natural scenes that combine ambient light coming from both the left or right side of the subjects. This reduces the strong backlighting and forces the illuminance distribution of the captured faces to be more diverse in the WDR scenes.\\
\textbf{Take the photographs.}
The subjects were asked to stand close to the source of ambient light (like a glass wall) that enabled a significant amount of ambient light, and look straight forward at the center camera. In each scene, we randomly chose participants to take one-face or multi-face (2,3,4 subjects) photos, preventing human factors from interfering with experimental results. We also increased the number of faces in each WDR scene to create different lighting intensity on each face. This is done by requiring multiple subjects to appear in different positions where the lighting intensity on their faces varies between each other. The maximum number of faces in a WDR scene is four. In this scenario, the influence of the flashlight is significantly smaller compared to the high intensity of the ambient light. Figure \ref{f-6e648c21ecc1} depicts the source of light that generates the WDR.\\
\textbf{Clean data.}
Due to limitations of the equipment and the shooting environment, the dynamic range of the captured images differs from time to time. A calculation to the dynamic range of all the captured images is required for removing the images with insufficient dynamic range. We excluded the images with a dynamic range of less than 32,768:1. Additionally, we performed face detection on all the images using different face detectors in this experiment. If there is a face that can't be detected by any face detectors, we consider this face undetectable, and exclude it. We label the coordinates of all faces based on the detection results.
\bgroup
\fixFloatSize{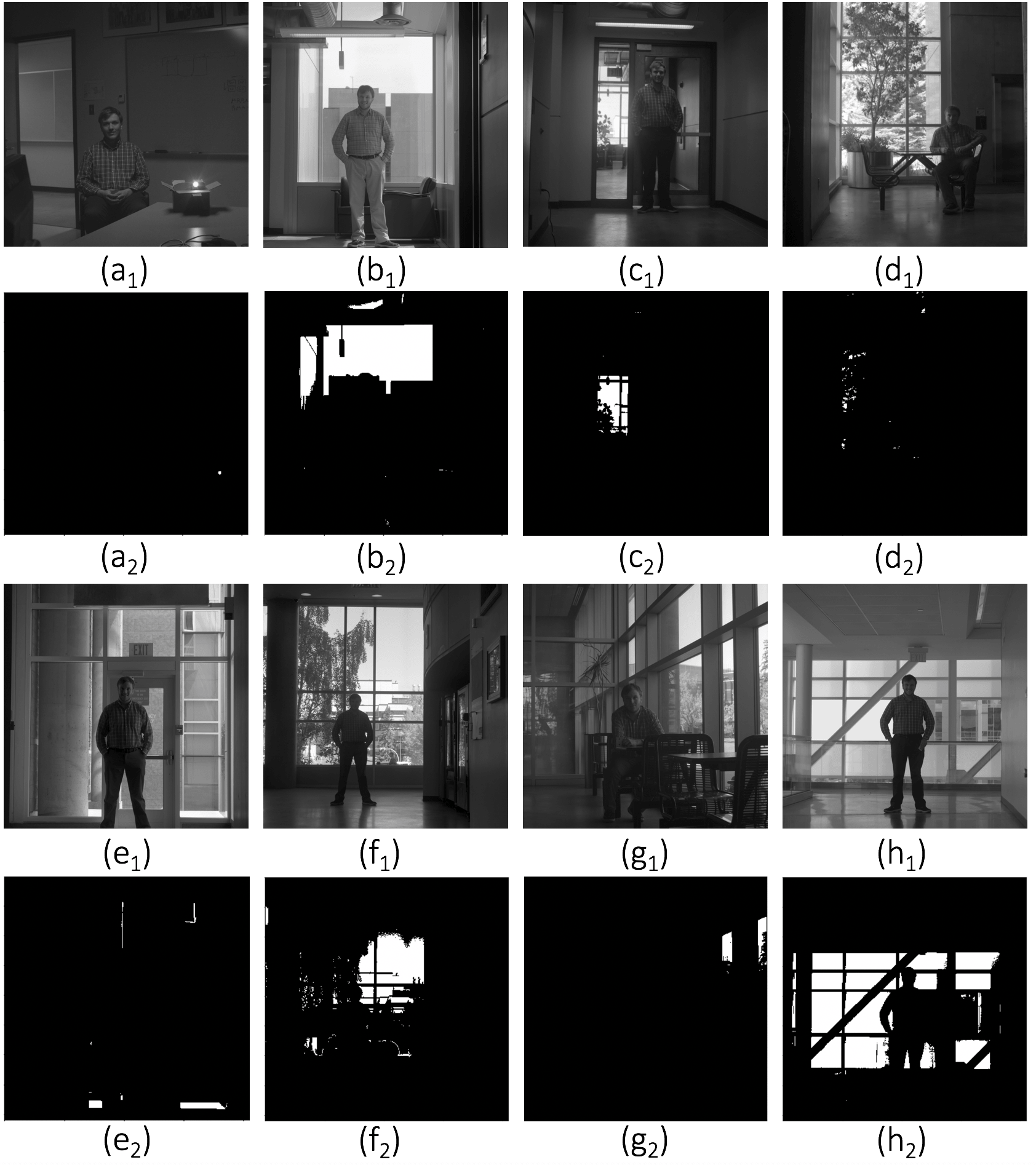}
\begin{figure*}[!htbp]
\centering \makeatletter\IfFileExists{images/4.png}{\includegraphics{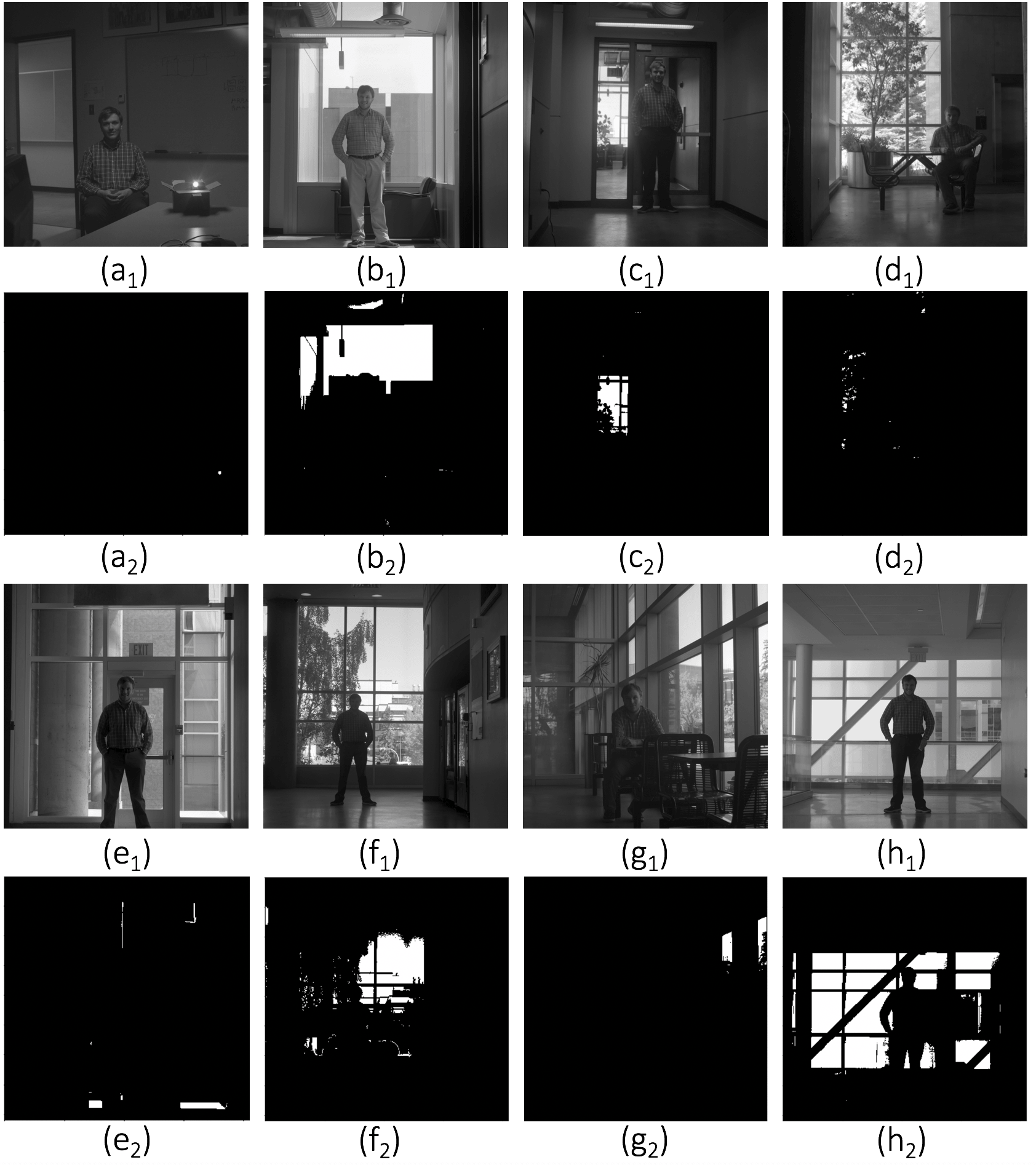}}{}
\makeatother 
\caption{{Demonstration of the WDR image captured from each scene.  The light source of an image and the binary masked representations. ($a_1$) The WDR image captured in the established artificial Scene.   ($b_1$) - ($h_1$) The WDRIs captured in the seven different natural scenes.  ($a_2$) The binary masked image of ($a_1$) with the threshold of 32,768:1. Only the point light source reached WDR.  ($b_2$) - ($h_2$) The binary masked image of ($b_1$) - ($h_1$), respectively, with the threshold of 32,768:1.  The white region shows the light source which established the WDR scenes.}}
\label{f-6e648c21ecc1}
\end{figure*}
\egroup

\subsection{Properties of WDR FACE}
\textbf{Variation.}
As stated in the first section, our focus is not on creating a powerful face detector, but on the effect of dynamic range on face detection. If we also diversify our database to a high degree of variability such as scale, pose, occlusion, expression, appearance, etc., it will complicate the detection results, making it more difficult to analyze.  For example, if we ask the subject to wear a mask, leading to inability to detect this face,  we won't be able to distinguish whether it is caused by facial occlusion or the WDR of the original image. To reduce the uncertainty, we narrowed the range of variation by requiring the subjects to not cover their faces (except for glasses) and to face the camera. We also tried to keep the scale of each face.  Such unification will help WDR FACE focus on the effect of dynamic range of images. The dynamic range of 398 images reached 32,768:1.  Among them, the dynamic range of 90\% images surpasses 60,000:1, and that of 70\% images exceed 65,000:1. Except for these 398 images, we got 12 images that have a dynamic range between (16,383:1, 32,768:1].  And the dynamic range of another three images is less than 16,383:1. The dynamic range of these three images is 16,143:1, 16,354:1, and 15,536:1, respectively.\\ 
\textbf{Size.}
Our database contains 398 WDR megapixel images from 29 subjects, distributed over eight WDR scenes.  The resolution of each image is 2048 x 2048. The size of each image is about 17MB, making the total size of WDR FACE about 7GB. To avoid overfitting, future researchers could perform data augmentation to increase the size of our dataset. Specifically, can do pre-processing like: image rotation, image shearing, image scaling, image blurring, image stretching, image translation and create at least 200 images per scene. Training neural network on a small dataset is a trend in the recent machine learning field, our dataset could accelerate tiny machine learning \citep{905417:20518687} of face detection on mobile devices. Besides, transfer learning \citep{905417:20518688} can be adopted to solve the face detection problem under our dataset, which does not require a large amount of training data.\\
\textbf{Format.}
The original image format is SIF by the Andor Zyla PLUS camera. Since this format is not commonly used, we converted it to the lossless TIFF, which is the format of our released database image.  All images are in a grayscale of 4.2 megapixels.  Since most of TMOs do not support grayscale format, we duplicated the grayscale to be 3-channel to feed in TMOs.  The TM results contain three channels with the same value.  We then extract one channel of the pixel value out and send it as input for the face detection stage.

\section{Evaluation Methodology}
\label{sect:eval_method} 
\subsection{Selection of Operators}
\subsubsection{Tone Mapping Operators Selection}
The role of a TMO is to map the WDRIs to the LDR domain while preserving the feature. Different TMOs have corresponding advantages in achieving the tasks.  We employed 25 different types of TMOs from various tools including Luminance HDR \footnote{https://github.com/LuminanceHDR/LuminanceHDR} and HDR toolbox \citep{905417:20518677}.  These two commonly used software packages provide a good sample, of fairly used tone mapping algorithms, where the code was easily ready.  They are the works of Mantiuk08 \citep{905417:20518642}, Fattal \citep{905417:20518669}, Ferradans \citep{905417:20518679}, Drago \citep{905417:20518637}, Durand \citep{905417:20518644}, Reinhard \citep{905417:20518639}, Photomatix \citep{905417:20518681}, BestExposureTMO\citep{905417:20518677}, ExponentialTMO\citep{905417:20518677}, LogarithmicTMO\citep{905417:20518677}, NormalizeTMO \citep{905417:20518677}, and of Ashikhmin \citep{905417:20518580}, Pattanaik \citep{905417:20518579}, Mai \citep{905417:20518680}, Ferwerda \citep{905417:20518578}, KimKautzConsistent \citep{905417:20518577}, Lischinski \citep{905417:20518576}, VanHateren \citep{905417:20518640}, Krawczyk \citep{905417:20518570}, Kuang \citep{905417:20518575}, Schlick \citep{905417:20518574}, Tumblin \citep{905417:20518573}, WardGlobal \citep{905417:20518675}, MSHIST \citep{905417:20518626}, and Gu \citep{905417:20518646},

\subsubsection{Face Detectors Selection}
As discussed in \textit{Section \ref{sect:related_work}}, there are many types of face detectors.  We chose 5 commonly used and publicly recognized high-performance face detectors in this experiment. They are Viola-Jones \citep{905417:20518591}, MMOD \citep {905417:20518587}, Faster R-CNN \citep{905417:20518584}, MTCNN \citep{905417:20518585}, and YOLOv3 \citep{905417:20518589}.  There were 3 different Haar Cascade Classifiers used for our experiment in the Viola-Jones (VJ) method.  They are \textit{alt}, \textit{alt\_tree}, and \textit{default}.  We used AlexNet and VGG16 as the backbone of Faster R-CNN (F-R-CNN) which were made for the ImageNet Challenge.  MMOD was made by using HoG features in conjunction with structural SVMs and it is available in the Dlib open-source library \citep{905417:20518581}.

\subsection{Procedure and Settings}
We first mapped the prepared 398 WDRIs to the LDR domain through the 25 different TMOs. After this step, 25 sets of LDR representations of WDRIs were generated, in total $25\times398 = 9,950$ tone-mapped copies.  Then, we fed these images as the input to 8 different face detectors.  The total number of results is $8\times25\times398 = 79,600$. After the experiment, we summarized all results and finally performed a descriptive analysis.  Figure \ref{f-754b6d748bcc} shows the pipeline of the experimental processes.
\bgroup
\fixFloatSize{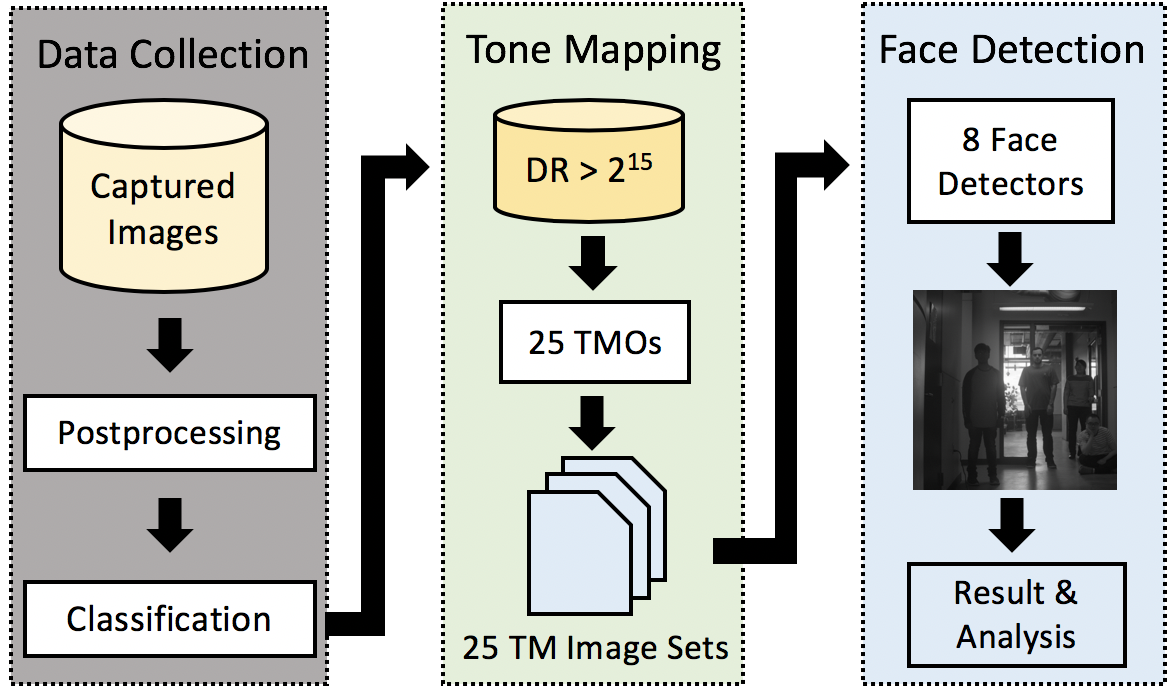}
\begin{figure*}[!htbp]
\centering \makeatletter\IfFileExists{images/5.png}{\includegraphics{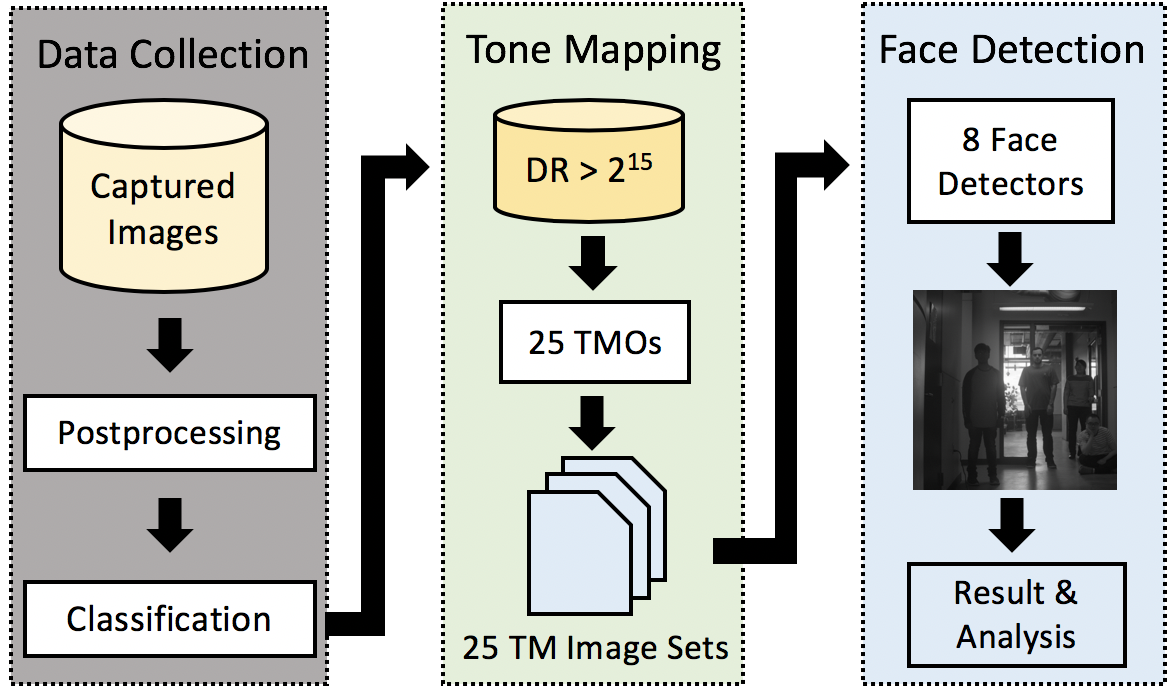}}{}
\makeatother 
\caption{{The pipeline of data collection and experiment.}}
\label{f-754b6d748bcc}
\end{figure*}
\egroup

\begin{table*}[!t]
\footnotesize
\caption{ The accuracy of face detection result generated by tone mapping after Gu \citep{905417:20518646} based on WDR scenes. Column 1 of WDR Scenes indicates the artificial WDR Scene and the rest are natural scenes.  The cells containing highest accuracy across all scenes are highlighted in bold. The w.ave. represents the weighted average accuracy by the percentage of the number of images in each scene.  The balance represents the averaged accuracy across all scenes where each scene is weighted equally.}
\label{table:fdor_vs_scene}
\tabcolsep 9.5pt 
\setlength{\tabcolsep}{4pt}
\begin{tabular*}{\textwidth}
{@{\extracolsep{\fill}}lcccccccccc}
\toprule
Accuracy (\%) & \multicolumn{10}{c}{WDR Scenes} \\
\midrule
Face Detectors & 1 & 2 & 3 & 4 & 5 & 6 & 7 & 8 & w.ave. & balance \\
\midrule
VJ\_alt & \textbf{44.81} & 26.51 & 29.55 & 9.57 & 24.32 & 16.99 & 19.22 & 20.88 & 24.74 & 23.98 \\
VJ\_alt\_tree & \textbf{44.81} & 26.19 & 29.55 & 9.57 & 24.32 & 16.99 & 19.22 & 20.88 & 24.69 & 23.94 \\
VJ\_default & \textbf{44.96} & 30.56 & 30.14 & 10.32 & 26.94 & 16.55 & 20.90 & 26.24 & 26.58 & 25.83 \\
MMOD & 78.38 & 68.91 & 58.06 & 46.08 & 54.65 & 30.30 & 50.59 & \textbf{84.78} & 60.23 & 58.97 \\
F-R-CNN A. & 51.28 & 63.08 & 38.76 & 46.58 & 60.22 & 12.33 & 70.41 & \textbf{89.36} & 54.23 & 54.00 \\
F-R-CNN V. & 69.83 & 71.19 & 53.91 & 73.58 & 73.33 & 35.14 & 80.43 & \textbf{90.32} & 68.67 & 68.47 \\
MTCNN & \textbf{98.20} & 72.88 & 88.80 & 87.16 & 87.36 & 61.84 & 87.36 & 93.62 & 85.39 & 84.65 \\
YOLOv3 & \textbf{99.10} & 75.21 & 94.44 & 95.19 & 98.82 & 82.61 & 95.40 & 97.85 & 92.35 & 92.33 \\
\bottomrule
\end{tabular*}
\end{table*}

\subsection{Metrics}
We computed the accuracy of face detection by the equation:
\begin{equation}
    Accuracy = (TP + TN) / (TP + FP + TN + FN) 
\end{equation}
where TP, FP, TN, and FN represent true positive, false positive, true negative, and false negative, respectively.  TP is the number of boxes that contain an intersection over union value 0.5, that is the predicted bounded box overlaps with the ground truth box and that the region of overlap exceeds 50\%. FP is the number of boxes that contain an intersection over union value $\leq 0.5$. TN is the number of images that does not contain faces and no faces are detected (our TN is zero since every image contains one or more faces). FN is the number of faces in the image that are not detected.

\section{Results and Discussion}
\label{sect:result} 
The accuracy of the detection result and analysis in the image set with the dynamic range \textgreater 32,768:1 is provided for an overview of the performance of different face detectors on the tone mapped WDRIs. The WDR face images can avoid pixel saturation compared with LDR images, which include a much higher level of detail and are closer to the range of human vision. Therefore, detection in the WDRIs would be much easier.\\
\textbf{Accuracy.} The details of the detection result by TMOs are shown in Table \ref{table:tmo_vs_fdor}. We highlight both the highest detection accuracy in each TMO and the cells that contain the highest accuracy across all TMOs. Each face detector performs differently in each tone-mapped LDR image set. YOLOv3 achieved the highest detection accuracy in all TMOs. Its best accuracy is 92.49\% achieved in the LDR image set generated via WardGlobal.  In terms of the tone mapped LDR images, the LDR image sets generated by Gu and WardGlobal are the most friendly TMOs to face detectors.  Four face detectors (MMOD, Fast R-CNN Alexnet, Fast R-CNN VGG16, and MTCNN) achieved the highest accuracy in the LDR image set tone mapped by Gu. The remaining face detectors (Viola-Jones alt, alt\_tree, default, and YOLOv3) reached the best accuracy in the LDR image set tone mapped by WardGlobal.  The most difficult LDR image set for face detection is generated by Schlick.  All face detectors except YOLOv3 remain under 1\% accuracy in this image set.  Faster R-CNN Alexnet wasn't able to correctly detect any faces in this image set.\\
\textbf{Gu TMO.} From another perspective, we show in Table \ref{table:fdor_vs_scene} the highest average accuracy in terms of scenes generated by Gu tone mapping operator \citep{905417:20518646} which is the TMO with the highest average accuracy. This table depicts the difficulty of face detection for each scene. We take the percentage of the number of images in each scene as the weight and calculated the weighted average.  The weighted average and balance metrics show that the scenes with higher weights generally have higher accuracy between 0.02\% to 1.26\%. Unsurprisingly, five TMOs (Viola-Jones alt, alt\_tree, default, MTCNN, and YOLOv3) achieve the best face detection accuracy in the artificial Scene (Scene 1) against the natural scenes because of the less complex illuminance distribution. The remaining TMOs (MMOD, FastR-CNN Alexnet, and Fast R-CNN VGG16) achieve the best accuracy in Scene 8. Scene 6 is the most challenging one where the ambient light was overwhelmed by the severe backlighting, resulting in the subjects' faces showing only silhouettes of heads.  

\section{Conclusion and Future Work}
\label{sect:conclusion} 
In this paper, we proposed WDR FACE, the first WDR database for face detection research. WDR FACE contains a total of 398 16-bit megapixel WDR image taken in eight selected WDR scenes.  The dynamic range of 90\% images surpasses 60,000:1, and that of 70\% images exceed 65,000:1.  We have provided a very detailed description of the WDR FACE. Furthermore, We conducted a face detection experiment using 25 different TMOs and five different face detectors. We provided a preliminary experimental face detection result to the unique WDR face database.
In summary, the contribution of this work is as follows:
\begin{enumerate}
   \item We have explained the issues of face detection under WDR scenes and we propose the first WDR database to support the study in this area.   The database contains 29 subjects for a total of 398 16-bit megapixel WDRIs. We have provided a very detailed description of this unique WDR database including the coordinates of the face bounding boxes.  
   \item We conducted face detection experiments using 25 different TMOs (see Figure \ref{f-c48060293e1a}) and five different face detectors.  We provide this preliminary face detection results on our unique WDR face database. This can serve as a basis for new TMO and face detection development evaluation in WDR imaging.
\end{enumerate}

Future work will focus on many potential facial identification applications with the WDR FACE database.  Preliminary results can serve as a basis for new TMO and face detection development evaluation in WDR imaging.  In addition to developing a new TMO or a face detector operates on LDR images,  we also consider extending our work to propose a novel face detection method that can skip the biases produced from the TM step and directly operates on WDRIs uses this database.

\section*{Acknowledgements}We thank all volunteers who participated in the image collection. We thank Andor Technology for the hardware support. This work was funded by Alberta Innovates (AI) Integrated Intelligent Sensors program (grant number RT735246), and by Natural Sciences and Engineering Research Council of Canada (NSERC)(grant numbers RT731684 and 10023039). We gratefully acknowledge the support of Dr. Yanuskevich. 
    \pdfbookmark[section]{References}{references}

\bibliographystyle{emerald-bib}

\bibliography{article}

\section{Appendices}
\label{sec14}
This Appendix contains the detailed results for the accuracy of each face detector for each tone mapping operator (Tables \ref{table:tmo_vs_fdor}).

\begin{table*}[b]
\footnotesize
\caption{The accuracy of each face detector for each tone mapping operator (unit in \%).  The name of 25 TMOs (summarized in \textit{Selection of Operators} section) used are noted in each column. Ashi., Ferr., Ferw., KimK., Kraw., Patta., Rein., VanHa., W.Global., B., E., N., L., VJ, F-R-CNN A., and F-R-CNN V. refer to Ashikhmin\citep{905417:20518580}, Ferradans\citep{905417:20518679}, Ferwerda\citep{905417:20518578}, KimKautz\citep{905417:20518577}, Krawczyk\citep{905417:20518570}, Pattanaik\citep{905417:20518579}, Reinhard\citep{905417:20518639}, VanHateren\citep{905417:20518640}, WardGlobal\citep{905417:20518675}, BestExposureTMO\citep{905417:20518677}, ExponentialTMO\citep{905417:20518677}, NormalizeTMO\citep{905417:20518677}, LogarithmicTMO\citep{905417:20518677}, Viola-Jones\citep{905417:20518591}, Faster-R-CNN Alexnet\citep{905417:20518584}, and Faster-R-CNN VGG16 respectively. The highest detection accuracy in each TMO is highlighted in bold. The cells containing highest accuracy across all scenes are highlighted in underline.}
\label{table:tmo_vs_fdor}
\tabcolsep 8pt 
{\begin{tabular*}{\textwidth}
{@{\extracolsep{\fill}}lccccccc}
\toprule
Accuracy (\%) & \multicolumn{7}{c}{Tone Mapping Operators} \\
\midrule
Detectors & Ashi. & B.TMO & Drago & Durand & E.TMO & Fattal & Ferr. \\ \midrule
VJ\_alt & 11.90 & 14.60 & 19.45 & 19.11 & 22.99 & 19.29 & 22.15 \\
VJ\_alt\_tree & 11.90 & 14.60 & 19.44 & 19.09 & 22.95 & 19.26 & 22.13 \\
VJ\_default& 11.18 & 13.86 & 20.12 & 19.54 & 24.56 & 20.87 & 23.66 \\
MMOD & 34.14 & 20.96 & 37.53 & 35.21 & 52.23 & 41.74 & 52.28 \\
F-R-CNN A. & 9.54 & 21.09 & 35.22 & 38.53 & 43.42 & 44.34 & 53.40 \\
F-R-CNN V. & 14.64 & 36.98 & 48.21 & 47.34 & 40.53 & 56.23 & 53.76 \\
MTCNN & 5.06 & 38.49 & 54.44 & 58.12 & 73.61 & 66.42 & 82.21 \\
YOLOv3 & \textbf{75.41} & \textbf{68.87} & \textbf{86.84} & \textbf{85.79} & \textbf{92.24} & \textbf{89.21} & \textbf{92.39} \\
\midrule
\midrule
Detectors & Ferw. & Gu & KimK. & Kraw. & Kuang & Lischinski & L.TMO \\
\midrule
VJ\_alt & 4.70 & 20.85 & 17.77 & 12.95 & 16.76 & 15.17 & 19.61 \\
VJ\_alt\_tree & 4.70 & 20.83 & 17.75 & 12.94 & 16.74 & 15.16 & 19.57 \\
VJ\_default& 3.37 & 22.73 & 16.96 & 10.85 & 16.24 & 13.95 & 19.53 \\
MMOD & 2.72 & \underline{60.64} & 27.51 & 19.41 & 24.26 & 21.52 & 26.57 \\
F-R-CNN A. & 0.90 & \underline{54.32} & 23.39 & 17.04 & 23.84 & 13.67 & 2.71 \\
F-R-CNN V. & 6.45 & \underline{68.79} & 39.19 & 37.85 & 41.11 & 28.95 & 6.37 \\
MTCNN & 7.04 & \underline{85.25} & 47.95 & 39.47 & 46.56 & 37.10 & 14.77 \\
YOLOv3 & \textbf{22.51} & \textbf{92.30} & \textbf{69.25} & \textbf{68.65} & \textbf{66.16} & \textbf{62.55} & \textbf{76.15} \\
\midrule
\midrule
Detectors & Mai & Mantiuk & MSHIST & N.TMO & Patta. & PhotoMatix & Rein. \\
\midrule
VJ\_alt & 21.51 & 21.88 & 21.00 & 12.98 & 4.34 & 20.11 & 18.33 \\
VJ\_alt\_tree & 21.49 & 21.85 & 20.99 & 12.97 & 4.34 & 20.09 & 18.31 \\
VJ\_default& 23.22 & 22.64 & 22.04 & 12.41 & 3.34 & 21.41 & 18.62 \\
MMOD & 51.28 & 40.85 & 48.34 & 14.62 & 6.11 & 43.97 & 35.30 \\
F-R-CNN A. & 51.83 & 42.79 & 51.34 & 11.93 & 0.90 & 46.92 & 28.61 \\
F-R-CNN V. & 50.00 & 51.88 & 57.90 & 20.21 & 18.45 & 55.20 & 50.19 \\
MTCNN & 78.09 & 66.12 & 77.43 & 25.64 & 14.99 & 67.55 & 52.66 \\
YOLOv3 & \textbf{92.13} & \textbf{90.61} & \textbf{91.15} & \textbf{45.01} & \textbf{30.48} & \textbf{90.87} & \textbf{83.46}  \\
\midrule
\midrule
Detectors & Schlick & Tumblin & VanHa. & W.Global & - & - & -  \\ 
\midrule
VJ\_alt & 0.71 & 15.56 & 2.28 & \underline{29.09} & - & - & -  \\
VJ\_alt\_tree & 0.71 & 15.55 & 2.28 & \underline{29.02} & - & - & -  \\
VJ\_default& 0.19 & 14.09 & 1.85 & \underline{31.25} & - & - & -  \\
MMOD & 0.13 & 22.45 & 14.53 & 50.75 & - & - & -  \\
F-R-CNN A. & 0.00 & 19.54 & 0.00 & 42.92 & - & - & -  \\
F-R-CNN V. & 0.65 & 39.19 & 2.47 & 44.57 & - & - & -  \\
MTCNN & 0.13 & 39.02 & 2.60 & 78.00 & - & - & -  \\
YOLOv3 & \textbf{4.13} & \textbf{67.97} & \textbf{34.41} & \underline{\textbf{92.49}} & - & - & -  \\
\bottomrule
\end{tabular*}}{}
\end{table*}

\end{document}